\pdfoutput=1

\documentclass[11pt]{article}

\usepackage[final]{acl}
\usepackage{times}
\usepackage{latexsym}
\usepackage{graphicx}
\usepackage{comment}
\usepackage{url}
\usepackage{lettrine}

\usepackage{multirow}
\usepackage{rotating}
\usepackage{booktabs}
\usepackage{times}
\usepackage{algorithm}
\usepackage{colortbl} 
\usepackage{amsmath}
\usepackage{algorithmic}
\usepackage{subcaption}
\usepackage{todonotes}
\usepackage{amssymb}

\usepackage[T1]{fontenc}

\usepackage[utf8]{inputenc}

\usepackage{microtype}

\usepackage{inconsolata}

\newcommand{\approachName}{SERPENT-VLM}
\title{\textbf{\textit{\approachName}} \includegraphics[scale = 0.13]{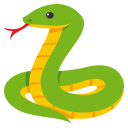} : Self-Refining Radiology Report Generation Using Vision Language Models 
}

\author{Manav Nitin Kapadnis\footnotemark[1]$\thanks{~~Equal contribution.}$  ~~~ Sohan Patnaik\footnotemark[1] ~~~ Abhilash Nandy  ~~~ Sourjyadip Ray \\ \textbf{Pawan Goyal} ~~~ \textbf{Debdoot Sheet} \\ \texttt{iammanavk@gmail.com ~~~ sohanpatnaik106@gmail.com} \\ Indian Institute of Technology Kharagpur \\ India}

\begin{document}
\maketitle
\begin{abstract}
\textit{Radiology Report Generation} (R2Gen) demonstrates how Multi-modal Large Language Models (MLLMs) can automate the creation of accurate and coherent radiological reports. Existing methods often \textit{hallucinate} details in text-based reports that don't accurately reflect the image content. To mitigate this, we introduce a novel strategy, \textbf{\textit{\approachName}}\ (\textbf{\underline{SE}}lf \textbf{\underline{R}}efining Radiology Re\textbf{\underline{P}}ort G\textbf{\underline{EN}}era\textbf{\underline{T}}ion using \textbf{\underline{V}}ision \textbf{\underline{L}}anguage \textbf{\underline{M}}odels), which improves the R2Gen task by integrating a self-refining mechanism into the MLLM framework. We employ a unique \textit{self-supervised loss} that leverages similarity between pooled image representations and the contextual representations of the generated radiological text, alongside the standard Causal Language Modeling objective, to refine image-text representations. This allows the model to scrutinize and align the generated text through dynamic interaction between a given image and the generated text, therefore reducing hallucination and continuously enhancing nuanced report generation. \approachName\ outperforms existing baselines such as LlaVA-Med, BiomedGPT, etc., achieving SoTA performance on the IU X-ray and Radiology Objects in COntext (ROCO) datasets, and also proves to be robust against noisy images. A qualitative case study emphasizes the significant advancements towards more sophisticated MLLM frameworks for R2Gen, opening paths for further research into self-supervised refinement in the medical imaging domain.
\end{abstract}


\section{Introduction} 

\label{Chapter 1} 

\textit{Radiology Report Generation} (R2Gen) serves as a crucial link between medical imaging and natural language processing, to automate the interpretation of radiological images into comprehensive text reports. This task requires models to learn long-range dependencies effectively while generating the report, a challenge that remains largely unmet in current systems. The primary goal of R2Gen is to generate accurate and comprehensive medical reports from radiological imagery, an essential step toward enhancing diagnostic accuracy and efficiency.
\begin{figure*}[!hbtp]
    \centering
    \includegraphics[width = 0.95\textwidth]{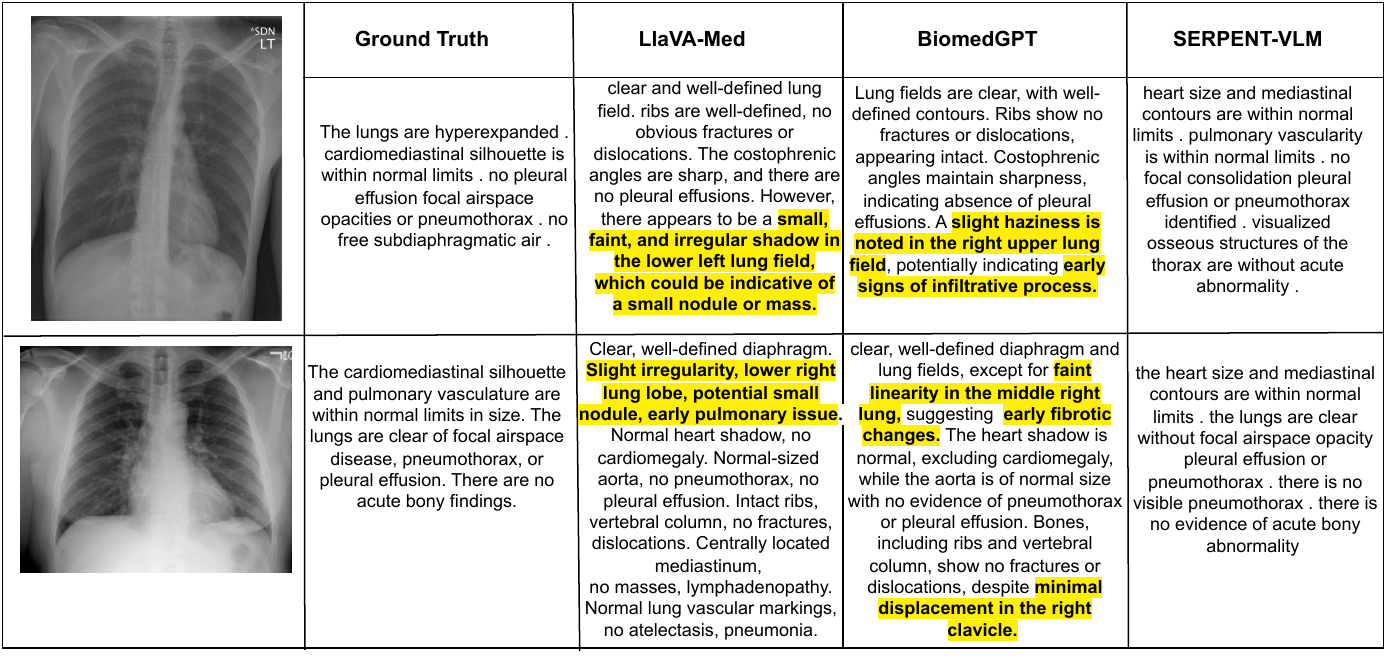}
    \caption{Generated report samples on IU-Xray dataset. We qualitatively analyze reports generated by medical pre-trained LLMs LlaVA-Med and BioMedGPT with \approachName. Hallucinated information in the reports is highlighted using yellow.}
    \label{fig:hallucination}
\end{figure*}
Prevailing methods \cite{vinyals2015showtell,xu2016showattendtell,stay_in_grid,image_captioning_attention,tang2021clip4caption} in R2Gen often rely on (1) large datasets for pre-training to impart domain-specific knowledge, and (2) typically utilizing compute-intensive encoder-decoder architectures for fine-tuning. These approaches are fraught with drawbacks, such as omission of minor yet clinically significant details \cite{Automatic_RRG_transformer,you2022aligntransformer,paper_6_18} and the persistent issue of \textit{hallucination} as seen in Fig. \ref{fig:hallucination}, where generated reports from LlaVA-Med and BiomedGPT wrongly include details not present in the images. Minimizing hallucinations in radiology report generation is crucial since these inaccuracies can lead to misdiagnoses, directly impacting patient treatment plans and outcomes. Moreover, reducing hallucinations ensures the reliability and trustworthiness of automated reports, which is vital for maintaining clinical credibility and facilitating effective patient care. Therefore, the limitations pertaining to existing approaches underscore the necessity for a more refined approach for accurate medical diagnosis, addressing the critical gaps in R2Gen. 

In this paper, we introduce a streamlined pipeline, \approachName, which begins by processing a given X-ray image by passing it through a visual encoder and mapping it to a vector representation in a high-dimensional space. This process facilitates a nuanced understanding of the medical imagery. The encoded image, alongside a report generation prompt, is then passed as inputs to a Large Language Model (LLM) for text generation. We employ a cross-entropy loss for the \textbf{causal language modeling objective} and \textbf{introduce a novel self-refining objective} that leverages the pooled image representation and the generated report's contextual representation. This allows for tuning the network without compromising inference latency, while significantly improving performance evaluated using metrics such as $Bleu$, $Rouge_L$, $BertScore$.

The contributions of our work are summarized as follows:
\begin{enumerate}
    \item Our approach \textbf{does not compromise on inference latency}, adopting a refining strategy through a novel loss function used only for fine-tuning
    \item The introduction of a self-refining loss ensures the generation of nuanced, \textbf{hallucination-free} radiology reports
    \item Our system not only matches but surpasses the performance of leading generalistic pre-trained medical LLMs.
    \item Our approach demonstrates \textbf{robustness against noisy image} inputs, maintaining the generation of comprehensive reports.
\end{enumerate}   

This marks a substantial advancement in the field of R2Gen, setting new benchmarks for accuracy, efficiency, and robustness.

The remainder of the paper is organized as follows: We begin by delving into the literature review in Section \ref{Chapter 2}, focusing on current and past state-of-the-art (SoTA) methodologies in the domain of radiological report generation. Section \ref{Chapter 4} discusses the proposed strategy for the self-refining fine-tuning our approach. The datasets, baselines, experimental setups, and ablation studies are detailed in Section \ref{Chapter 6_experimental}. Finally, we conclude with a summary of our findings in Section \ref{Chapter 7}.
\section{Related Work} 

\label{Chapter 2} 

\textbf{Medical Report Generation (MRG)}:
Medical Report Generation has been extensively studied through ML models. \cite{paper_6_15} proposed a co-attention network that aligns visual and textual information to generate comprehensive radiology reports. Further enhancing the capabilities, a memory-driven transformer \cite{chen-etal-2020-generating} integrates memory modules for encoding and decoding processes, allowing for more sophisticated report generation \cite{chen-etal-2020-generating,chen2022crossmodal}. Cross-modal learning \cite{wang2022crossmodal} utilizes prototype matrices and contrastive losses to refine the learning of visual-textual correlations, complemented by a self-boosting framework to align image features with report text \cite{paper_6_18}. \cite{radreport_ieee} addressed the problem of mitigating inherent biases through a data-driven method, introducing a prior-posterior knowledge-based report generation. \cite{paper_6_20} leveraged curriculum learning to extract global concepts to create a bridge between images and text. Task-specific architecture with sentence-level attention mechanism across visual features \cite{paper_6_23} allows the model to capture key medical concepts from images. A weakly supervised paradigm to amplify hard negative samples \cite{paper_6_22} addresses the medical data scarcity challenge.
\vspace{1mm}

\noindent\textbf{Large Language Models and Vision language Models}: The advent of Large Language Models (LLMs) such as GPT-4, Claude, BARD showcase excellent zero-shot language understanding \cite{brown2020language,li2021xmodaler,radreport_ieee,irvin2019chexpert}; image understanding and visual question answering \cite{team2023gemini} capabilities.
Open-source LLMs, like LLaMA and BLOOM, and Multi-modal LLMs such as LlaVA \cite{liu2024visual}, Open Flamingo \cite{awadalla2023openflamingo} have also democratized access to cutting-edge generative technology \cite{ouyang2022training,pan2020xlinear}. 
Furthermore, domain-specific models LlaVA-Med \cite{li2023llavamed} and BiomedGPT \cite{zhang2024biomedgpt} have shown promising results in pathology and radiology-related tasks. However, knowledge grounding for medical reports \cite{hyland2023maira}, thereby reducing hallucination produced by these models remains a challenge.
\vspace{1mm}

\noindent\textbf{Source \& Representation of Feedback}: Iterative refinement in MRG has traditionally relied on human feedback to achieve high-quality outputs \cite{tandon2022learning}. Scalar reward functions and domain-specific feedback tools, such as compilers, were proposed as cost-effective alternatives to human feedback \cite{le2022coderl,yasunaga2020graphbased}. Recent developments show that Large Language Models (LLMs) can self-evaluate their responses. However, applying this to Multi-modal Large Language Models remains largely unexplored in terms of generating grounded and hallucination-free responses.

We now discuss the proposed methodology in the subsequent section.
 

\section{Methodology} 

\label{Chapter 4} 

\begin{figure*}[h]
    \centering
    \includegraphics[width = \textwidth]{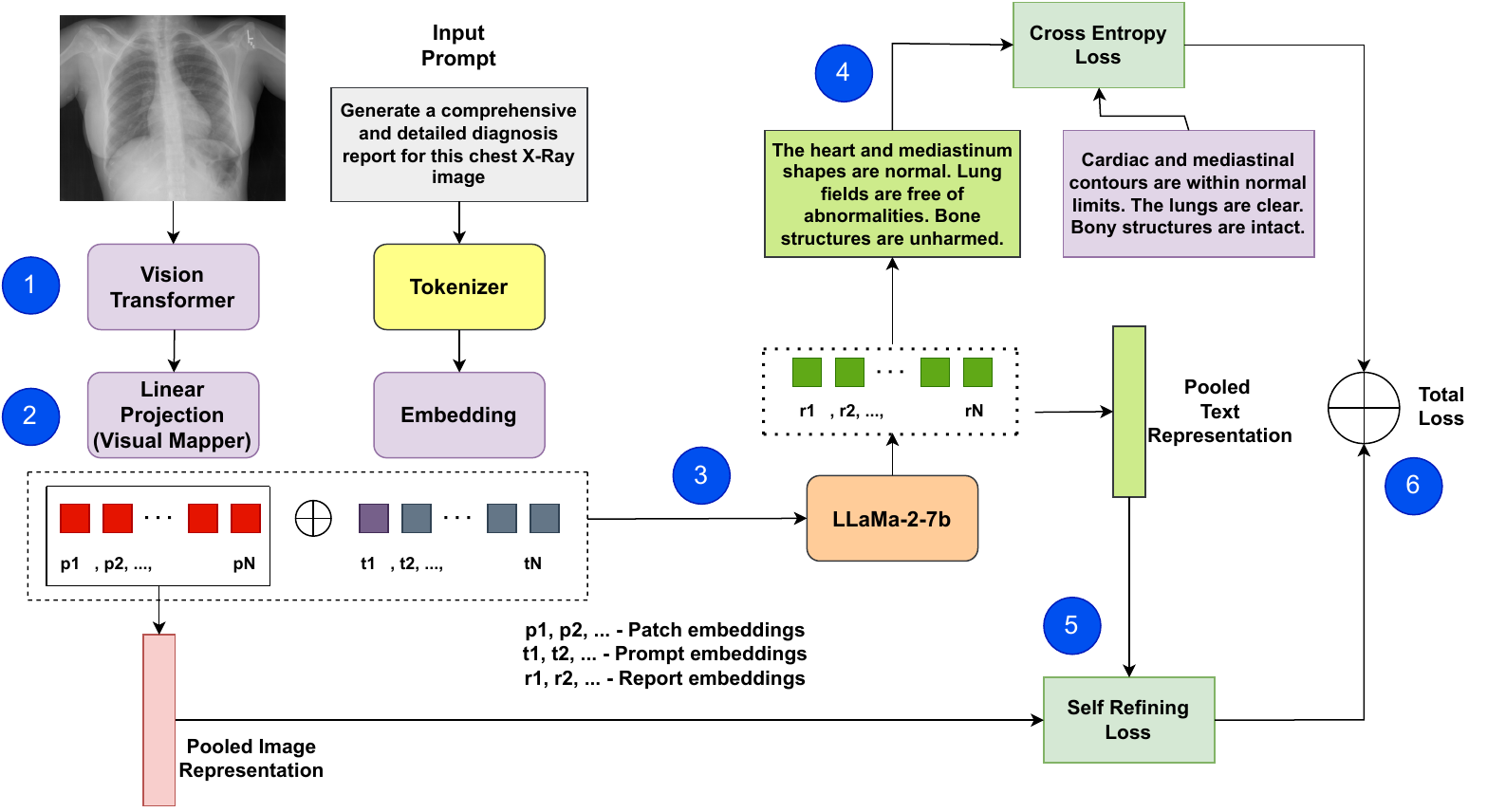}
    \caption{Overview of the \approachName\ pipeline. The X-ray image is processed using a visual encoder (step 1) and projected onto a high-dimensional space using a visual mapper (step 2). The encoded image with the report generation prompt is fed into the LLM (step 3). Cross-entropy loss is employed (step 4) for the causal language modeling objective. The pooled image representation and the Contextual representation of the generated report are used to compute the self-refining loss (step 5). A weighted combination of both objectives is used to train the network (step 6).}
    \label{fig:overview}
\end{figure*}
\subsection{Overview of \approachName}
We summarize the pipeline of \approachName\ in Figure \ref{fig:overview}. It consists of two branches to establish the learning optimization criterion. \textbf{1) Causal Language Modeling Objective} enforces standard cross-entropy loss (step 4 in Fig. \ref{fig:overview}) for supervised radiology report generation. Our approach consists of a visual encoder that extracts information from chest X-ray images (step 1 in Fig. \ref{fig:overview}), a visual mapper that projects low dimensional image features onto high dimensional feature space (step 2 in Fig. \ref{fig:overview}) and a Large Language Model that autoregressively generates the diagnostic radiological report (step 3 in Fig. \ref{fig:overview}). To further reduce hallucination, we construct a pooled representation of the given X-ray image, a contextual representation leveraging the attention weights and last hidden states of the generated report and enforce \textbf{2) Self Refining Objective} that tries to maximise the similarity between pooled image representation and the contextual representation of the generated report through a self-supervised loss criterion (step 5 in Fig. \ref{fig:overview}). We train the network through a weighted combination of both the losses (step 6 in Fig. \ref{fig:overview}), thereby enabling \approachName\ to continuously refine itself by aligning generated text with the input image. We now discuss the details of each component.

\subsection{\approachName\ Framework}

The architecture of \approachName\ can be partitioned into three different modules - a visual encoder, a visual mapper and a large language model (LLM). Formally, consider a chest X-ray image $I_v \in \mathbb{R}^{C\text{x}H\text{x}W}$, where $C$ is the number of input channels, $H$, $W$ being the height and width of the image respectively. $I_v = [I_{v_1}, I_{v_2}, \cdots I_{v_k}]$ comprises of a sequence of $k$ patches with $I_{v_i} \in \mathbb{R}^{C \text{x} P \text{x} P}$ being the $i^{th}$ patch, and $P$ is the patch size. We leverage a transformer-based visual encoder $V_{enc}$ to encode and obtain contextual representation $\Tilde{e}_{v_i} \in \mathbb{R}^{d_v}$ denoted by Eq. \ref{Eqn:image_encode} and aggregate each encoded patch to obtain a global image representation $\Tilde{e}_v$ depicted by Eq. \ref{Eqn:pooled_image_rep}. 
\begin{equation}
    \Tilde{e}_{v_1}, \Tilde{e}_{v_2}, \cdots \Tilde{e}_{v_k} = V_{enc}(I_{v_1}, I_{v_2}, \cdots I_{v_k})
    \label{Eqn:image_encode}
\end{equation}
\begin{equation}
    \Tilde{e}_v = V_{pooler}(\Tilde{e}_{v_1}, \Tilde{e}_{v_2}, \cdots \Tilde{e}_{v_k})
    \label{Eqn:pooled_image_rep}
\end{equation}
The encoded image features inherently reside in a visual feature space, which is distinct and not directly compatible with the textual feature space, and hence need to be aligned with the word embedding space of the LLM. To ensure this, we use a learnable visual mapper $V_{map}$ to project the patch embeddings $\Tilde{e}_{v_i}$ onto the word embedding space. Formally, $e_{v_i} = V_{map}(\Tilde{e}_{v_i})$. We construct a seed prompt $T$ instructing the LLM to generate a report conditioned on the image $I_v$, and obtain the corresponding tokens $\mathcal{T}_{tokens} = [t_1, t_2, \cdots, t_{|\mathcal{T}_{tokens}|}]$ which is given as input to the $Embedding$ module of the LLM to construct the token embeddings (refer Eq. \ref{Eqn:token_embeddings}), 
\begin{equation}
\resizebox{\columnwidth}{!}{$
    e_{t_1}, e_{t_2}, \cdots, e_{t_{|\mathcal{T}_{tokens}|}} = Embedding(t_1, t_2, \cdots, t_{|\mathcal{T}_{tokens}|})
$}
\label{Eqn:token_embeddings}
\end{equation}

We concatenate the sequence of projected image patch embeddings $e_{v_i}$ with the seed prompt text embeddings $e_{t_j}$ to obtain a sequence of input embeddings $e_{\mathcal{I}} = [e_v; e_t]$ which are given as input to the decoder-only LLM denoted by \textbf{$TD$} for generating the logits of the response tokens in auto-regressive fashion. $V_{enc}$, $V_{pooler}$, $V_{map}$ and $TD$ are trained through cross-entropy loss $\mathcal{L}_{report}$ enforced between the generated logits and the actual responses. To further guide the report generation process by aligning the generated response with the input image, we enforce a self-supervised \textit{refining loss}. 

\subsection{Self-refining Strategy}
\label{sec:self-refine}
We construct an aggregated representation of the generated text by utilizing the attention weights of the last layer of $TD$. Consider the logit distribution for each generated token as $l_i \in \mathbb{R}^{d}$, where $d$ is the vocabulary size of $TD$. To encode the representation of each generated token, which is further used to compute the \textit{self-refining} loss in a differentiable fashion, we leverage Gumbel-Softmax on the logit distribution to obtain $\hat{l}_i$ for each predicted token. We construct the aggregated representation $\hat{e}_i^p = \sum_{j=1}^d e_j \hat{l}_{ij} $ of each predicted token by taking a weighted sum of the embedding matrix $E = e_1, e_2, \cdots, e_d$ with $\hat{l}_i$ being the corresponding weights. Formally, 
\begin{equation}
    \hat{l}_{ij} = \frac{e^{(log(l_{ij}) + g_{ij}) / \tau}}{\sum_{j=1}^d e^{(log(l_{ij}) + g_{ij}) / \tau}}
\end{equation}
Since, the gumbel-softmax operator makes the logit distribution peaky, taking a weighted sum effectively yields the predicted token embeddings. Further, we construct an aggregated representation $h_t \in \mathbb{R}^{d_t}$ of the predicted token embeddings by leveraging the attention weights from the last layer of $TD$. We hypothesize that aligning the aggregated representation of the generated report with the pooled input image representation would reduce hallucination and ground the report generation task. For this, we enforce a \textit{self-refining loss} between $h_t$ and $e_v$ depicted by Eq. \ref{Eqn:refine_loss}
\begin{equation}
    \mathcal{L}_{refine} = \frac{1}{b} \sum_i^b e^{-h_t^Te_v},
\end{equation}
where \( b \) is the batch size.
\label{Eqn:refine_loss}

Minimizing the negative exponential of the similarity between the image and generated text representation pushes the representation closer, thus further grounding the report generation process. We optimize our network with a weighted combination of both the causal language modeling objective and the self-refining objective. The total loss is denoted by Eq. \ref{Eqn:total_loss}
\begin{equation}
    \mathcal{L}_{total} = \lambda_{report} \text{ } \mathcal{L}_{report} + \lambda_{refine} \text{ } \mathcal{L}_{refine}
    \label{Eqn:total_loss}
\end{equation} 
$\mathcal{L}_{report}$ depicts the standard causal language modeling objective that ensures the conditional generation of radiological report text based on the input image, whereas $\mathcal{L}_{refine}$ ensures that the generated report is grounded in context of the input image, thereby establishing a robust pipeline for radiology report generation. 

\section{Experiments and Evaluation} 

\label{Chapter 6_experimental} 


We now discuss the details corresponding to the experiments and ablation studies carried out and enumerate the observations.

\subsection{Implementation Details} 
We discuss the technical details and hyper-parameter settings for all the experiments. For the visual encoder $V_{enc}$, we employed the base version of Swin-Transformer-V2\footnote{https://huggingface.co/microsoft/swinv2-base-patch4-window12-192-22k} and a feed-forward neural network for $V_{map}$. We leverage LLaMA2-7B\footnote{https://huggingface.co/meta-llama/Llama-2-7b-chat-hf} as our primary LLM. Further, the hidden dimension of $d_v$ of $V_{enc}$ and $d_t$ of $TD$ are $768$ and $1024$ respectively. We freeze the weights of $V_{enc}$, however keep $V_{map}$ trainable. We employ LoRA with a rank and $\alpha$-scaling factor of 16 each to fine-tune the underlying LLM $TD$. We train \approachName\ for 15 epochs on IU-Xray dataset and 20 epochs on the ROCO dataset with mixed precision on an effective batch size (BS) of 6 using one NVIDIA A40 48GB GPU using a learning rate of $1\times10^{-4}$ with linear rate scheduler through AdamW optimizer. For inference, we leverage beam search decoding with beam size configured to 3.
\vspace{1mm}

\subsection{Datasets and Evaluation Metrics:} 
We evaluate \approachName\ on two commonly used datasets diverse modality - 
\begin{enumerate}
    \item \textbf{IU X-Ray} which is a widely used publicly available dataset for medical report generation tasks containing 3,955 fully de-identified radiology reports with sections such as Impression, Findings, Indication, etc., each associated with frontal and/or lateral chest X-rays, totaling 7,470 images;
    \item \textbf{ROCO} which has `radiology' and `out-of-class' subsets (synthetic radiology images, clinical photos, portraits, compound radiology images, and digital art) of roughly 65,460 and 8,182 `radiology', and 4,902 and 613 `out-of-class' images in the train and test set respectively.
\end{enumerate}
Since the reports are verbose and need to be accurately measured with word-level precision, we compute overlap-based metrics like BLEU and Rouge-L, and a semantic similarity-based metric BertScore for evaluating the efficacy of our approach. 

\begin{table}[htbp]
\begin{center}
\resizebox{0.9\columnwidth}{!}{\begin{tabular}{@{}lllll@{}}

\toprule
\textbf{Dataset} & \textbf{Train} & \textbf{Val} & \textbf{Test} & \textbf{Image Views} \\ \midrule
IU X-Ray         & 2769           & 791          & 395           & Frontal and Lateral  \\
ROCO             & 65460          & 8183         & 8182          & Frontal              \\ \bottomrule
\end{tabular}}
\end{center}
\caption{Statistics of Evaluation Datasets}
\label{tab:dataset_stats}
\end{table}

\subsection{Performance of \approachName\ on Radiology Report Generation}
\begin{table*}[htbp]
\resizebox{2\columnwidth}{!}{\begin{tabular}{@{}l|cccccc|cccccc@{}}
\toprule
\multicolumn{7}{c|}{\textbf{IU-Xray}} &
  \multicolumn{6}{c}{\textbf{ROCO}} \\ \midrule
\textbf{Methods} &
  \textbf{$Bleu_1$} &
  \textbf{$Bleu_2$} &
  \textbf{$Bleu_3$} &
  \textbf{$Bleu_4$} &
  \textbf{$Rouge_L$} &
  \textbf{$BertScore$} &
  \textbf{$Bleu_1$} &
  \textbf{$Bleu_2$} &
  \textbf{$Bleu_3$} &
  \textbf{$Bleu_4$} &
  \textbf{$Rouge_L$} &
  \textbf{$BertScore$} \\\hline
Show-Tell &
  0.243 &
  \cellcolor[HTML]{FFFFFF}{\color[HTML]{1F1F1F} 0.13} &
  0.108 &
  \cellcolor[HTML]{FFFFFF}{\color[HTML]{1F1F1F} 0.078} &
  0.307 &
  0.378 &
  0.104 &
  0.076 &
  0.051 &
  0.027 &
  0.089 &
  0.34 \\
Att2in &
  0.248 &
  0.134 &
  0.116 &
  0.091 &
  0.309 &
  0.386 &
  0.106 &
  0.077 &
  0.052 &
  0.027 &
  0.091 &
  0.347 \\
AdaAtt &
  0.284 &
  0.207 &
  0.15 &
  0.126 &
  0.311 &
  0.442 &
  0.122 &
  0.089 &
  0.060 &
  0.031 &
  0.104 &
  0.397 \\
Transformer &
  0.372 &
  0.251 &
  0.147 &
  0.136 &
  0.317 &
  0.579 &
  0.159 &
  0.116 &
  0.079 &
  0.041 &
  0.137 &
  0.521 \\
M2transformer &
  0.402 &
  0.284 &
  0.168 &
  0.143 &
  0.328 &
  0.626 &
  0.172 &
  0.125 &
  0.085 &
  0.044 &
  0.148 &
  0.563 \\
R2Gen &
  0.47 &
  0.304 &
  0.219 &
  0.165 &
  0.371 &
  0.732 &
  0.201 &
  0.147 &
  0.099 &
  0.052 &
  0.173 &
  0.658 \\
R2GenCMN &
  0.475 &
  0.309 &
  0.222 &
  0.17 &
  0.375 &
  0.74 &
  0.169 &
  0.148 &
  0.100 &
  0.052 &
  0.175 &
  0.665 \\
MSAT &
  0.481 &
  0.316 &
  0.226 &
  0.171 &
  0.372 &
  0.749 &
  0.212 &
  0.150 &
  0.102 &
  0.053 &
  0.177 &
  0.673 \\
METransformer &
  0.483 &
  0.322 &
  0.228 &
  0.172 &
  0.38 &
  0.752 &
  0.211 &
  0.151 &
  0.102 &
  0.053 &
  0.178 &
  0.676 \\
R2GenGPT (Deep) &
  0.480 &
  0.316 &
  0.216 &
  0.169 &
  0.377 &
  0.748 &
  0.213 &
  0.150 &
  0.101 &
  0.053 &
  0.177 &
  0.672 \\
MiniGPT4 &
  0.494 &
  0.329 &
  0.220 &
  0.179 &
  0.390 &
  0.767 &
  0.219 &
  0.156 &
  0.103 &
  0.056 &
  0.183 &
  0.689 \\
BiomedGPT &
  0.516 &
  0.343 &
  0.233 &
  0.183 &
  0.403 &
  0.793 &
  0.229 &
  0.163 &
  0.109 &
  0.058 &
  0.189 &
  0.712 \\
LlaVA-Med &
  0.528 &
  0.346 &
  0.237 &
  0.186 &
  0.422 &
  0.845 &
  0.234 &
  0.164 &
  \textbf{0.111} &
  \textbf{0.061} &
  0.198 &
  0.759 \\
SERPENT-VLM &
  \textbf{0.547} &
  \textbf{0.356} &
  \textbf{0.242} &
  \textbf{0.190} &
  \textbf{0.452} &
  \textbf{0.935} &
  \textbf{0.243} &
  \textbf{0.169} &
  0.108 &
  0.057 &
  \textbf{0.212} &
  \textbf{0.84} \\ \bottomrule
\end{tabular}}
\caption{Results of SERPENT-VLM on Benchmark datasets}
\label{tab:results_table}
\end{table*}

Table \ref{tab:results_table} illustrates the comprehensive comparison of SERPENT-VLM against various state-of-the-art baselines across the IU-Xray and ROCO datasets. In comparison with traditional non-LLM approaches such as  Show-Tell \cite{vinyals2015showtell}, Att2in \cite{xu2016showattendtell}, and R2Gen \cite{chen-etal-2020-generating}, \approachName\ exhibits significant improvements. For instance, on the IU-Xray dataset, SERPENT-VLM achieves a $Bleu_4$ score of 0.190, surpassing Show-Tell's 0.078 and R2Gen's 0.165, and even outperforming the more advanced R2GenCMN, which scores 0.170. This indicates not only an improvement in capturing long-range dependencies but also a notable reduction in detail hallucination, a common issue in earlier models. Furthermore, when compared to Medical LLMs and generalistic Vision-Language Models such as LlaVA-Med \cite{li2023llavamed}, BiomedGPT \cite{zhang2024biomedgpt}, and MiniGPT4 \cite{zhu2023minigpt4}, \approachName\ demonstrates superior performance, marking a significant leap in R2Gen. For example, against LlaVA-Med, which records a $Bleu_4$ of 0.186 on IU-Xray, SERPENT-VLM shows a marked improvement with a score of 0.190. Similarly, in the context of $BertScore$, SERPENT-VLM achieves an impressive 0.935 compared to LlaVA-Med’s 0.845 and BiomedGPT’s 0.793, underscoring its enhanced textual coherence.

\subsection{Discussion on the Impact of different Design Choices for \approachName}
We carry experiments pertaining to two different design choices for \approachName\ and establish the efficacy of the proposed architecture through the comparative analysis across experiments. 
\begin{enumerate}
    \item \textbf{Effect of relative importance of two losses:} We vary the relative importance self-refining loss ($\lambda_{refine}$) and report-generation loss ($\lambda_{report}$) in Eq. \ref{Eqn:total_loss}. Table \ref{tab:lambda_ratio_table} shows that combining the two losses yields much better performance for IU X-ray and ROCO compared to just using the report generation loss (row 5 vs. row 2). This highlights that self-refining loss complements the report generation loss by grounding the generated report on the input image, thereby reducing hallucination. Further, it is observed that using only self-refining loss (row 1) leads to a degradation in performance because \approachName\ is trained only through a self-supervised paradigm without any kind of supervision. As observed, this equilibrium is not merely about avoiding hallucinations but also about fostering a synergistic effect where each loss component reinforces the other, thereby elevating the overall quality and reliability of the automated radiology reports. The findings from our experiments provide compelling evidence for the critical role of balanced loss parameters in achieving the desired outcomes, advocating for a nuanced approach in their application within the framework of \approachName.

    \begin{table*}[h]
\centering
\resizebox{1.5\columnwidth}{!}{ \begin{tabular}{@{}lcccccccc@{}}
\toprule
\textbf{Dataset} &
  \textbf{$\lambda_{Report}$} &
  \textbf{$\lambda_{Refine}$} &
  \textbf{$Bleu_1$} &
  \textbf{$Bleu_2$} &
  \textbf{$Bleu_3$} &
  \textbf{$Bleu_4$} &
  \textbf{$Rouge_L$} &
  \textbf{$BertScore$} \\ \midrule
\multirow{5}{*}{\textbf{IU-Xray}} & 0            & 1.0          & 0.416          & 0.270          & 0.184          & 0.144          & 0.344          & 0.711          \\
& \textbf{0.3} & \textbf{0.7} & \textbf{0.547} & \textbf{0.356} & \textbf{0.242} & \textbf{0.190} & \textbf{0.452} & \textbf{0.935} \\
& 0.5          & 0.5          & 0.492          & 0.320          & 0.218          & 0.171          & 0.407          & 0.842          \\
& 0.7          & 0.3          & 0.479          & 0.311          & 0.212          & 0.166          & 0.396          & 0.818          \\
& 1            & 0.0          & 0.451          & 0.311          & 0.200          & 0.157          & 0.373          & 0.771          \\ \midrule
\multirow{5}{*}{\textbf{ROCO}}    & 0            & 1            & 0.187          & 0.130          & 0.083          & 0.044          & 0.163          & 0.647          \\
  & \textbf{0.3} & \textbf{0.7} & \textbf{0.243} & \textbf{0.169} & \textbf{0.108} & \textbf{0.057} & \textbf{0.212} & \textbf{0.840} \\
  & 0.5          & 0.5          & 0.214          & 0.149          & 0.095          & 0.050          & 0.187          & 0.739          \\
  & 0.7          & 0.3          & 0.207          & 0.144          & 0.092          & 0.048          & 0.180          & 0.714          \\
  & 1            & 0            & 0.194          & 0.135          & 0.086          & 0.046          & 0.170          & 0.672          \\ \bottomrule
\end{tabular}}
\caption{Impact of combining self-refining loss (weight $\lambda_{refine}$) with report-generation loss (weight $\lambda_{report}$). Fusing both the loss components gives optimal performance.}
\label{tab:lambda_ratio_table}
\end{table*}

    \item \textbf{Effect of contextual representation design strategy:} We explore different aggregation strategies for obtaining the contextual representation of the generated report. As depicted in Table \ref{tab: aggregation technique results}, attention-based aggregation outperforms other aggregation strategies by a significant margin by obtaining a BertScore of 0.935 and 0.840; BLEU$_1$ score of 0.547 and 0.243 on IU X-ray and ROCO respectively. Average pooling (average of token representations), Max pooling (token representation with maximum L2-norm) and Top-k average pooling (average top $k=5$ token representations based on attention-weights) give sub-optimal performance on both IU X-ray and ROCO benchmark, thereby establishing the critical importance of sophisticated feature integration methods in enhancing the model's capability to synthesize coherent and contextually relevant radiology reports. Exploration into different aggregation strategies reveals that the sophistication and adaptability of the aggregation mechanism play a pivotal role in the efficacy of medical report generation models.
    \begin{table*}[htbp]
\centering
\resizebox{1.8\columnwidth}{!}{\begin{tabular}{@{}llcccccc@{}}
\toprule
\textbf{Dataset} &
  \textbf{Design Strategy} &
  \textbf{$Bleu_1$} &
  \textbf{$Bleu_2$} &
  \textbf{$Bleu_3$} &
  \textbf{$Bleu_4$} &
  \textbf{$Rouge_L$} &
  \textbf{$BertScore$} \\ \midrule
\multirow{4}{*}{\textbf{IU-Xray}} &
  \textbf{Attention based aggregation} &
  \textbf{0.547} &
  \textbf{0.356} &
  \textbf{0.242} &
  \textbf{0.190} &
  \textbf{0.452} &
  \textbf{0.935} \\
 & Average pooling                                  & 0.410 & 0.267 & 0.182 & 0.143 & 0.339 & 0.701 \\
 & Top k average pooling & 0.465 & 0.303 & 0.206 & 0.162 & 0.384 & 0.795 \\
 & Max pooling                                      & 0.383 & 0.249 & 0.169 & 0.133 & 0.316 & 0.655 \\ \midrule
\multirow{4}{*}{\textbf{ROCO}} &
  \textbf{Attention based aggregation} &
  \textbf{0.243} &
  \textbf{0.169} &
  \textbf{0.108} &
  \textbf{0.057} &
  \textbf{0.212} &
  \textbf{0.840} \\
 & Average pooling                                  & 0.190 & 0.132 & 0.084 & 0.044 & 0.165 & 0.655 \\
 & Top k average pooling & 0.199 & 0.139 & 0.089 & 0.047 & 0.174 & 0.689 \\
 & Max pooling                                      & 0.170 & 0.118 & 0.076 & 0.040 & 0.148 & 0.588 \\ \bottomrule
\end{tabular}}
\caption{Performance comparison of different design strategies for contextual representation. Attention weights-based aggregation displays superior performance.}
\label{tab: aggregation technique results}
\end{table*}

\end{enumerate}

\subsection{How robust is \approachName\ to noisy images?}
\begin{figure*}[htbp] 
    \centering
    \begin{subfigure}{\columnwidth} 
        \centering
        \includegraphics[width=\textwidth]{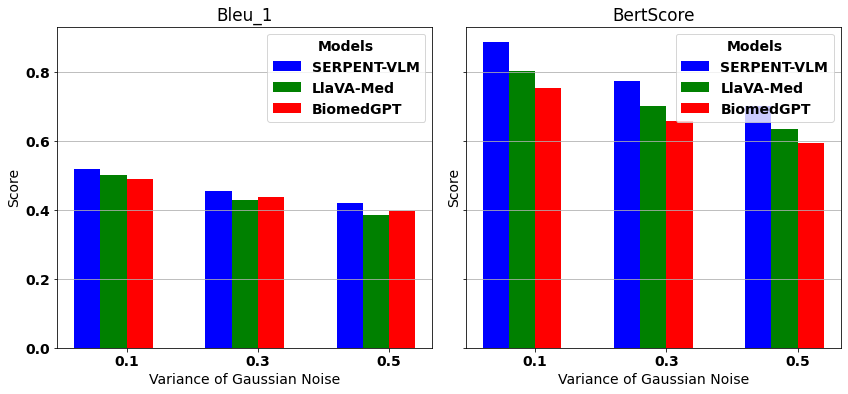}
        \caption{Performance metrics for ROCO dataset with varying levels of Gaussian noise added to input radiological images.}
        \label{subfig:roco_noise}
    \end{subfigure}
    \hfill 
    \begin{subfigure}{\columnwidth} 
        \centering
        \includegraphics[width=\textwidth]{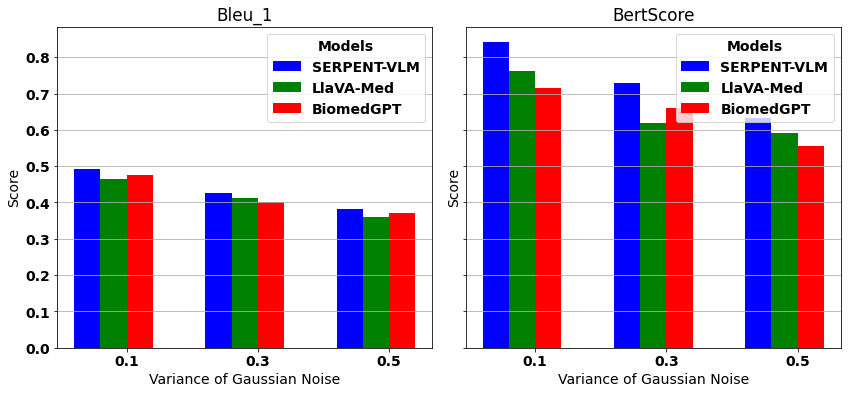}
        \caption{Performance metrics for IU-Xray dataset with varying levels of Gaussian noise added to input radiological images.}
        \label{subfig:iuxray_noise}
    \end{subfigure}
    \caption{Comparative performance metrics for ROCO and IU-Xray datasets.}
    \label{fig:datasets_noise}
\end{figure*}

We assess the robustness of SoTA methods LlaVA-Med and BiomedGPT, with our method \approachName, by introducing Gaussian noise to radiological images. Fig. \ref{fig:datasets_noise} demonstrate that \approachName\ significantly outperforms the current SoTA models, LlaVA-Med and BiomedGPT, across all Gaussian Noise scales, maintaining higher BLEU$_1$ (~5-6\% higher) and BertScore (~9-10\% higher) metrics, thus showcasing superior robustness in report generation under noisy and corrupted images. This also highlights \approachName's ability to focus on relevant parts of the image, thereby mitigating the effects of added noise and grounding the generated report - an indication of reduction in hallucination phenomena. The integration of \approachName\ could markedly enhance diagnostic accuracy, aiding radiologists in delivering faster and more accurate patient care.




\section{Summary and Conclusion} 
\label{Chapter 7} 

In this paper, we propose \approachName, an innovative method for producing detailed and accurate radiology reports from Chest X-rays without hallucinations. The process utilizes a frozen visual encoder to transform X-ray images into a high-dimensional space, which a Large Language Model (LLM) then uses to generate initial reports. These reports undergo further refinement through a novel combination of self-refining loss and Causal Language Modeling Loss, significantly surpassing existing methods as detailed in Section \ref{Chapter 6_experimental}. Our experiments in Section \ref{Chapter 6_experimental} and supplementary materials, confirm the effectiveness of our self-refining approach, even with distorted noisy images. Our future works involve the extension of our method to other medical imaging types, such as MRIs and CT scans, and to incorporate diagnostic RADreports to enhance report accuracy further.

\section*{Limitations}


The \approachName\ has shown significant advancements in creating radiology reports from chest X-rays, reducing inaccuracies, and better matching the content of the images compared to earlier models. However, this research has its limitations. The testing of the model's performance and adaptability has been limited to particular datasets (IU X-Ray and ROCO), which do not encompass the broad spectrum of radiological images or health conditions. It remains unclear how well this would work in actual medical situations. Furthermore, although the model's ability to handle low-quality images is emphasized, the wide range of image quality in real-life scenarios could pose challenges that have yet to be evaluated.

\section*{Ethics Statement}

The deployment of \approachName\ in clinical settings involves significant ethical considerations. The model’s potential to generate erroneous interpretations from radiological images, despite reduced hallucinations, necessitates cautious application, especially since incorrect reports could lead to misdiagnoses or inappropriate treatments. The use of large datasets for training also raises privacy concerns, requiring stringent data handling and patient consent protocols. 

\section*{Acknowledgements}

The project was supported in part by the grant given by I-Hub Foundation for Cobotics, IIT Delhi for the project, "Voice based Natural Interaction for Goal Oriented Tasks in Healthcare".

\bibliography{anthology,custom}




\end{document}